\documentclass[fleqn,10pt]{wlscirep}
\usepackage[utf8]{inputenc}
\usepackage[T1]{fontenc}

\title{Bringing Multi-Modal Multi-Task Federated Foundation Models to Education Domain: Prospects and Challenges}

\author[1]{Kasra Borazjani}
\author[2]{Naji Khosravan}
\author[3]{Rajeev Sahay}
\author[4]{Bita Akram}
\author[1]{\\ Seyyedali Hosseinalipour}
\affil[1]{University at Buffalo --SUNY, Department of Electrical Engineering, Buffalo, NY, USA}
\affil[2]{Adobe Research, Seattle, WA, USA}
\affil[3]{University of California -- San Diego, Department of Electrical and Computer Engineering, San Diego, CA, USA}
\affil[4]{NC State University, Department of Computer Science, Raleigh, NC, USA}

\affil[*]{Corresponding Author: S. Hosseinalipour (alipour@buffalo.edu)}

\begin{abstract}
Multi-modal multi-task (M3T) foundation models (FMs) have recently shown transformative potential in artificial intelligence, with emerging applications in education. However, their deployment in real-world educational settings is hindered by privacy regulations, data silos, and limited domain-specific data availability. We introduce M3T Federated Foundation Models (FedFMs) for education: a paradigm that integrates federated learning (FL) with M3T FMs to enable collaborative, privacy-preserving training across decentralized institutions while accommodating diverse modalities and tasks. Subsequently, this position paper aims to unveil M3T FedFMs as a promising yet underexplored approach to the education community, explore its potentials, and reveal its related future research directions.
We outline how M3T FedFMs can advance three critical pillars of next-generation intelligent education systems: \textit{(i) privacy preservation}, by keeping sensitive multi-modal student and institutional data local; \textit{(ii) personalization}, through modular architectures enabling tailored models for students, instructors, and institutions; and \textit{(iii) equity and inclusivity}, by facilitating participation from underrepresented and resource-constrained entities. We finally identify various open research challenges, including studying of
(i) inter-institution heterogeneous privacy regulations,
(ii) the non-uniformity of data modalities' characteristics,
(iii) the unlearning approaches for M3T FedFMs,
(iv) the continual learning frameworks for M3T FedFMs, 
and (v) M3T FedFM model interpretability, which must be collectively addressed for practical deployment.

\end{abstract}
\begin{document}

\flushbottom
\maketitle
%
%
\thispagestyle{empty}


\section*{Introduction}




The modern era has witnessed a surge in the use of artificial intelligence (AI) and machine learning (ML) to support a range of education-related tasks. These include predicting student learning outcomes and success~\cite{ofori2020using}, analyzing peer-to-peer collaboration patterns in online/in-person classrooms~\cite{hridi2025privacy}, monitoring students with behavioral or neurodevelopmental needs~\cite{barua2022artificial}, designing curricula for diverse educational settings~\cite{ball2019applying}, improving the students' mental health~\cite{ebrahimi2025transition}, and enabling personalized learning experiences in self-regulated learning environments~\cite{ingkavara2022use}.
With the expansion of AI/ML applications in education, two parallel trends have emerged. On one hand, leveraging multiple data modalities (e.g., text, audio, video, image) collected in educational environments to train \textit{multi-modal ML models}, capable of outperforming their uni-modal counterparts, has become a vibrant area of research~\cite{xie2025review, griol2014developing}. On the other hand, the use of these diverse modalities to train multi-task ML models that serve a variety of downstream educational tasks has also attracted growing attention~\cite{an2022no, geden2020predictive}. For example, video input in a humanoid robot can simultaneously support gesture tracking, object identification, and enhance speech understanding. As a result, the convergence of these trends has positioned \textit{multi-modal multi-task (M3T) learning} at the forefront of AI/ML applications in education~\cite{kuchemann2025opportunities, xu2024foundation}.

In parallel, the broader AI/ML community has undergone a significant transformation with the rise of M3T ML models. Initially popularized as foundation models (FMs) in the form of large language models (LLMs), such as GPT-3~\cite{gpt3}, BERT~\cite{devlin2019bert}, LLaMA~\cite{touvron2023llama}, and PaLM~\cite{chowdhery2023palm} focused primarily on text-based tasks, these models have recently evolved into M3T FMs, such as ChatGPT-4~\cite{openai2023gpt4}, Gemini~\cite{team2023gemini}, Llama-3~\cite{grattafiori2024llama}, and CLIP~\cite{guo2025human}. These emerging M3T FMs are capable of simultaneously processing multiple input modalities and capturing contextual relationships across multiple modalities and tasks. They have demonstrated remarkable generalization abilities, which is a result of (pre-)training on massive data.
Despite their promise, M3T FMs remain largely underexplored in the education domain mostly due to their recent emergence. In particular, the use of FMs in education has primarily focused on LLMs and text-based applications, such as automated feedback generation and essay grading~\cite{jia2024assessing}, question answering~\cite{mitra2024retllm}, and intelligent tutoring systems~\cite{molina2024leveraging}. Importantly, the influence of FMs and LLMs has extended beyond traditional classroom contexts. They are now being integrated into remote learning platforms, used to provide real-time feedback, and applied to educational research by offering analytical insights derived from classroom data~\cite{kuchemann2025opportunities}.

Although the use of M3T FMs in education has been proposed in a few recent studies~\cite{kuchemann2025opportunities,xu2024foundation}, a critical and largely unresolved question remains: \textit{Where does the data come from to train or fine-tune these data-hungry models in educational settings?} Specifically, educational tasks require domain-specific data, which is typically \textit{siloed} across multiple infrastructure layers, ranging from school-level and departmental servers to college and university data repositories. A major obstacle in utilizing this data lies in stringent data-sharing restrictions, including privacy regulations on both instutional and regional levels (e.g., FERPA) \cite{zeide2018learner}, ethical considerations, and student consent requirements~\cite{prinsloo2018student}, all of which prohibit the transfer of sensitive educational data to external servers for model training. 
As a result, the \textit{conventional centralized training/fine-tuning of M3T FMs} becomes infeasible for deploying them in real-world educational environments. Even if centralized access were possible to the above soiled data, the issue of data scarcity persists: high-quality, task-relevant educational data is often limited and fragmented across the isolated data sources. This challenge is further compounded by equity concerns, where models trained primarily on data from a single institution or demographically skewed population risk amplifying bias and marginalizing underrepresented or under-resourced learners.
Without addressing these fundamental barriers, the deployment of M3T FMs in education, despite their theoretical promise, remains largely aspirational.
In this paper, we propose a path forward by leveraging federated learning (FL)~\cite{mcmahan2017communication}, a pioneering distributed learning paradigm that enables collaborative model training without sharing raw data, for the training/fine-tuning of M3T FMs. Specifically, we give our perspective on \textit{M3T Federated Foundation Models (FedFMs) for education}, a novel direction that opens up an untapped research space at the intersection of M3T FMs, FL, and privacy-preserving human-centered AI/ML.

The remainder of the paper is organized as follows. We begin by reviewing the relevant literature on M3T FMs and FL within educational contexts. We then explore the potential of M3T FedFMs to advance education through three key dimensions: (1) privacy preservation, (2) personalization, and (3) equity enhancement. Finally, we discuss the key challenges associated with implementing M3T FedFMs in education and outline promising future research directions.

\section*{Overview on FL, M3T FMs, and M3T FedFMs} \label{sec:m3t-review}

~~~~\textbf{1. Federated Learning (FL):}
FL is a pioneering distributed ML paradigm that enables collaborative model training across multiple clients/participants (e.g., students, educators, institutions). FL operates through a series of global aggregation rounds, each comprising four key steps: (1) each  client  trains a local model on its own data (e.g., via stochastic gradient decent approach), (2) the locally trained models/gradients of clients are periodically sent to the server through  uplink transmissions, (3) the server aggregates (e.g., via weighted averaging) the received trained models to create an updated \textit{global model}, (4) the server broadcasts the updated global model to the clients, synchronizing their local models and initiating the next round of local model training. 
FL is widely regarded as a privacy-preserving distributed ML approach, as it replaces the transmission of sensitive raw data with model/gradient parameters. Note that although prior work has shown that even such transmitted parameters can are still prone to adversarial attacks, such as \textit{reconstruction attacks} that aim to regenerate training data~\cite{chen2022practical} or \textit{model inversion attacks} that extract client private information~\cite{li2022ressfl} from the transmitted parameters, several countermeasures exist,  including:
(1) \textit{Differential Privacy (DP)}~\cite{el2022differential}, which injects calibrated noise into transmitted parameters to obfuscate the underlying client data used to train the model, and
(2) \textit{Functional Encryption}~\cite{fang2021privacy, chang2023privacy}, where model parameters are encrypted in a way that allows only specific FL-related computations (e.g., model aggregation) to be performed without exposing the underlying client data.

By facilitating collaboration across a diverse network of institutions, FL helps overcome two key challenges typically faced when employing ML in education domains: \textit{(1) data scarcity,} by enabling isolated and limited datasets to contribute collectively to a shared global model, and \textit{(2) equity and inclusion,} by incorporating data from underrepresented or marginalized groups, distributed across different institutions, into the global model. 
Given these promising capabilities, FL has recently gained attention in the AI-assisted education literature~\cite{fachola2023federated, guo2020pedagogical,hridi2024revolutionizing,chu2022mitigating,chu2024multi}. Nevertheless, the majority of these studies focus on the adoption of FL for training of conventional ML models (e.g., convolutional neural networks) and have yet to explore the FL-driven training/fine-tuning of M3T FMs within the education domain.

\textbf{2. Multi-Modal Multi-Task Foundation Models (M3T FMs):}
M3T FMs are typically pre-trained on massive, heterogeneous datasets using self-supervised or unsupervised learning techniques, enabling them to acquire broad contextual understanding that can be effectively adapted to a wide range of domain-specific applications (e.g., enabling the operation of humanoid robots in domestic environments and extended reality systems)~\cite{borazjani2025multi, nadimi2025multi}. Fundamentally, M3T FMs extend the capabilities of conventional LLMs by incorporating multiple input modalities (e.g., text, audio, image, and video) and supporting a more diverse set of tasks (e.g., video understanding, conditional image generation, and image classification) alongside traditional text-based applications like question answering and text generation.  
Their potential in the education domain has recently been recognized~\cite{kuchemann2025opportunities}, with emerging use cases such as intelligent tutoring, automated feedback generation, and curriculum design. While the seminal work in~\cite{kuchemann2025opportunities} provides an in-depth analysis of the educational impact of M3T FMs, it does not address the specific training mechanisms behind these models and implicitly assumes their centralized training/fine-tuning. 
We therefore refer the reader to that work for broader context and position this paper as a complementary contribution, with the main focus on introducing the education community to the novelties of distributed, privacy-preserving, FL-driven training/fine-tuning of M3T FMs under the umbrella of M3T FedFMs.

\begin{figure}[t]
    \centering
    \includegraphics[width=\textwidth]{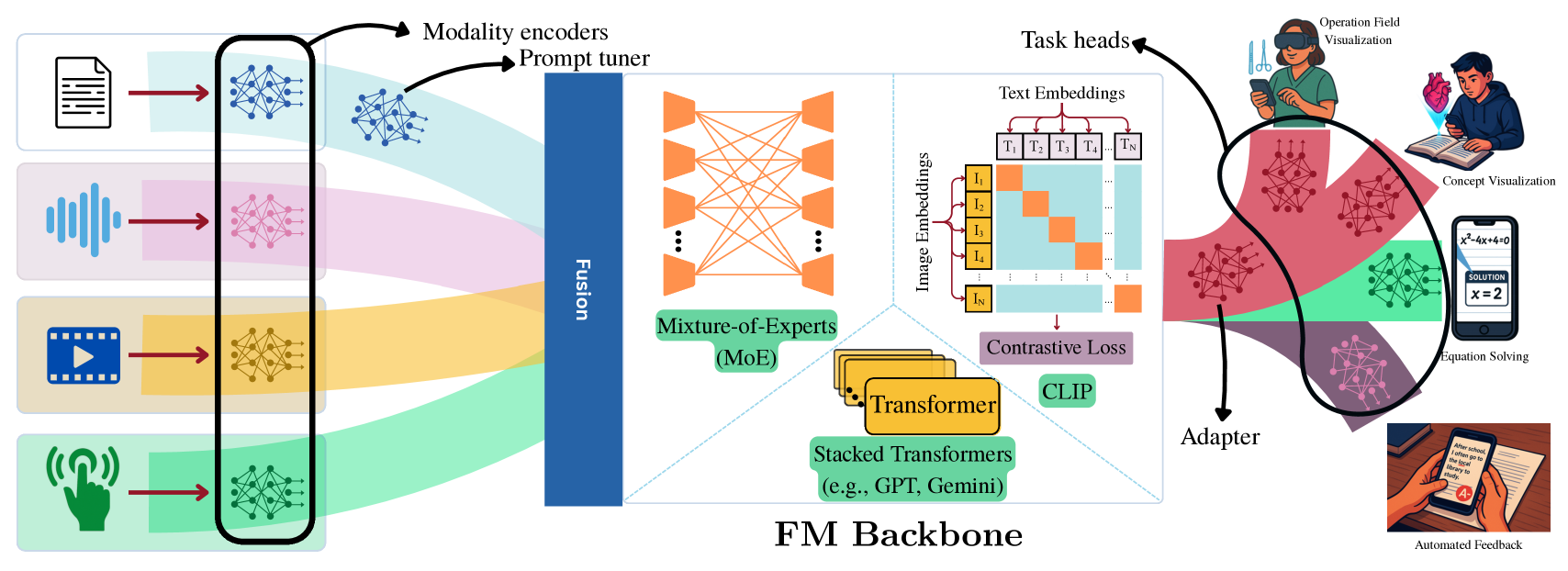}
       \vspace{-6mm}
    \caption{\small High-level architecture of an M3T FM, consisting of three major components. (1) \textit{Modality encoders:} These modules transform raw input data from various modalities into intermediate embeddings. Fusion of modality-specific embeddings can be achieved via simple concatenation or more sophisticated mechanisms such as neural fusion blocks or attention-based integration. It can also be non-existent in some cases (e.g., CLIP). (2) \textit{Backbone:} The backbone performs contextual reasoning, inter-modality correlation, and task generalization. It can be instantiated using various architectures, including Mixture-of-Experts (MoEs)\cite{chen2024disentanglement}, dual encoders (as in CLIP~\cite{guo2025human}), or stacked transformers (as in GPT models~\cite{openai2023gpt4}). (3) \textit{Task heads:} These are task-specific output layers that generate the results (e.g., classification labels, generated text) based on the representations produced by the backbone. M3T FMs also support lightweight fine-tuning strategies, where most of the model parameters are frozen and only a small subset is adapted. Three instances of these strategies are as follows. \textit{(1) Prompt tuners:}~\cite{guo2024scattering,jia2022visual} Modules that condition input embeddings to align with task-specific contexts. \textit{(2) Adapters:}~\cite{long2024dual,zhang2022contrastive} Trainable parameter blocks inserted at different depths of the model to enable rapid adaptation to new tasks or modalities. \textit{(3) Low-rank adaptations (LoRA):}~\cite{yang2024low,wen2023batched} Efficient fine-tuning methods that decompose and optimize a low-rank subset of parameters or adapter weights, significantly reducing training cost while preserving performance.}
    \label{fig:fm-architecture}
    \vspace{-5mm}
\end{figure}

To facilitate a clear understanding of their internal mechanisms, we next present a high-level overview of the general architecture of M3T FMs. 
Depicted in Fig. \ref{fig:fm-architecture}, M3T FMs architecturally consist of three main components: (1) \textit{modality encoders}, (2) \textit{backbone}, and (3) \textit{task heads}. They also can accommodate two additional components in their architecture, more commonly used in the scenarios entailing fine-tuning a pre-trained model to new contexts or tasks: \textit{prompt tuner}, and \textit{adapter}. For a more detailed description of these components refer to the caption/legend of Fig. \ref{fig:fm-architecture}. M3T FMs support a wide range of training regimes, offering flexibility to either \textit{train from scratch} or \textit{fine-tune on downstream tasks} after large-scale pretraining.

\textbf{3. Multi-Modal Multi-Task Federated Foundational Models (M3T FedFMs):}
M3T FedFMs can be understood as the FL-driven training of M3T FMs across a distributed set of clients. Similar to conventional FL, M3T FedFMs operate through a series of global aggregation rounds, each comprising the standard aforementioned four steps: (1) local training, (2) uplink transmission of local models/gradients, (3) server-side model aggregation, and (4) broadcast of the updated global model back to clients.
However, a key distinction between M3T FedFMs and traditional FL training of conventional ML models lies in the nature of local adaptation and aggregation. In M3T FedFMs, local training typically involves lightweight fine-tuning techniques, where only a subset of the model components, such as modality encoders, task heads, adapters, or prompt tuners, are updated. These components can then be selectively aggregated to produce a unified, fine-tuned global model that better generalizes across diverse client data distributions.
This modular\footnote{Here, ``modularity" refers to the capability of training various local M3T FM modules (e.g., encoders, task heads, adapters) independently across the clients.} training and aggregation approach~\cite{chen2024disentanglement} enables clients to obtain local M3T FMs suitable for their own tasks or modalities.


While M3T FedFMs hold great promise for enabling high-performance, locally adapted M3T FMs across distributed clients~\cite{chen2024disentanglement,chen2024feddat}, their implementation introduces a range of challenges. These include inherited issues from conventional FL (data heterogeneity~\cite{borazjani2025redefining,borazjani2024multi}, intermittent client connectivity~\cite{parasnis2023connectivity}, and limited client-side computational resources~\cite{chai2019towards}), as well as challenges specific to M3T FMs (e.g., selecting which components or parameters to fine-tune and aggregate). Moreover, the integration of M3T FMs with FL  introduces a set of unique challenges at their intersection (as will be explained later in the context of ``Challenges and Open Directions"), challenges that are not fully addressed by existing work in either field alone and are unique to M3T FedFMs. It is worth mentioning that M3T FedFMs represent a highly emerging topic within the AI/ML community, with only a handful of early studies exploring their theoretical foundations~\cite{chen2024feddat, chen2024disentanglement} and envisioning their applicability across domains such as healthcare~\cite{li2025open}, embodied AI~\cite{borazjani2025multi}, extended reality~\cite{nadimi2025multi}, and wireless edge/fog networks~\cite{abdisarabshali2025hierarchical}.
One promising domain still poised for breakthroughs enabled by M3T FedFMs is education, which we explore in the remainder of this paper to illuminate its unique applications and challenges.

\begin{figure}[!h]
    \centering
    \includegraphics[width=0.84\textwidth]{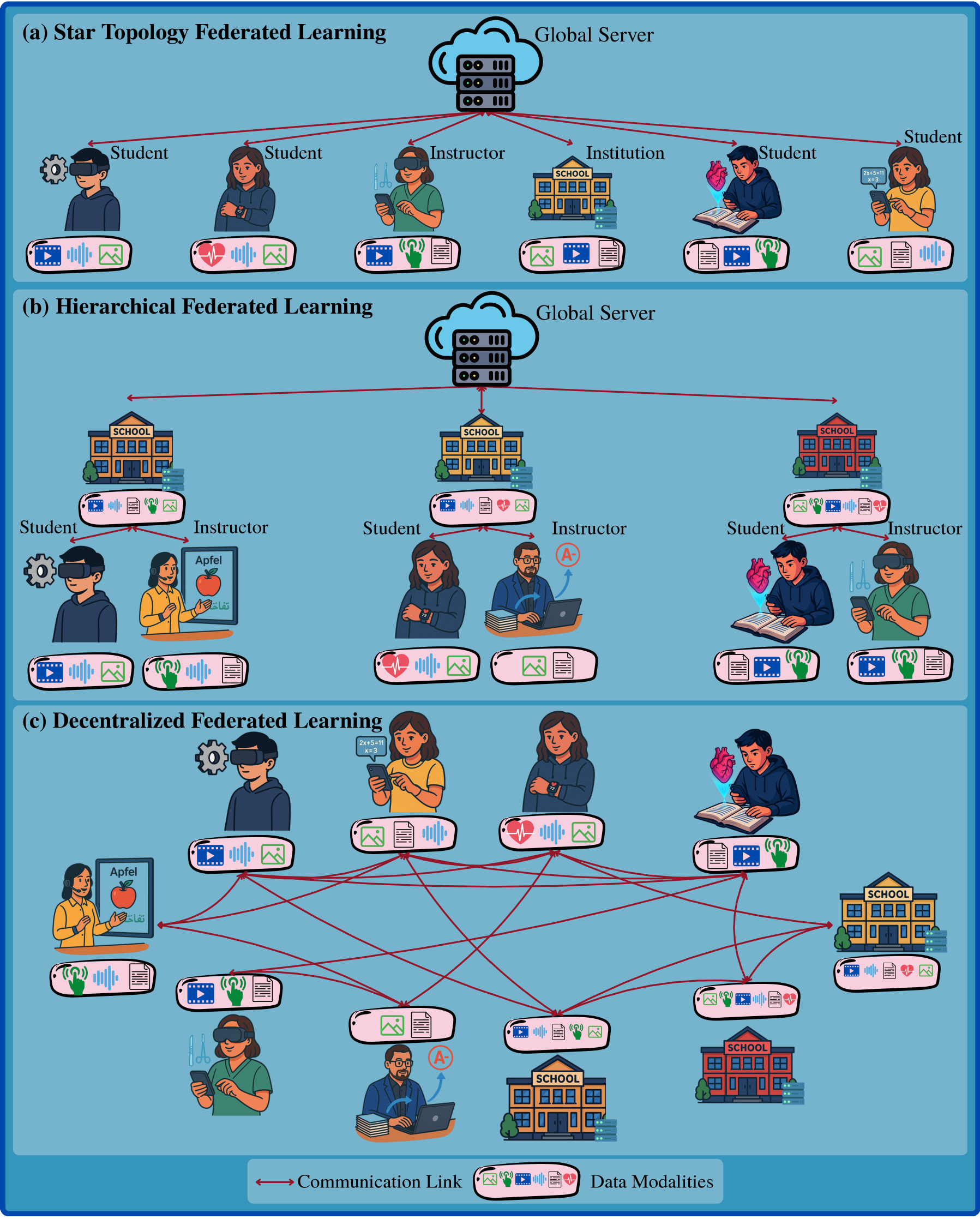}
    \vspace{-2mm}
    \caption{\small Various configurations can be adopted for M3T FedFM-enabled networks across clients in the education contexts,  three of which are depicted: \textbf{(a) Star Topology FL}: A global server maintains a comprehensive global model encompassing all task and modality variations present across the system. All clients (i.e., institutions, instructors, and students) are directly connected to the global server. Client receive  customized subsets of the global model tailored to their specific tasks and input modalities. Following local training/fine-tuning, clients updated models are transmitted back to the global server for aggregation.
    \textbf{(b) Hierarchical FL}:  The global server distributes task- and modality-relevant subsets of the global model to each institution. These subsets contain only the components required by the institution’s associated instructors and students. Each institution then relays the necessary parts of the model to its end users based on their individual task and modality needs. After local training/fine-tuning, local models of end users are sent back to the institutions for aggregation. These institution-level models can then be further aggregated at the global server, combining insights across multiple institutions.
    \textbf{(c) Decentralized FL}: Without a global server, model aggregation is performed in a decentralized manner across clients (e.g., through consensus-based methods). Clients exchange model updates directly with their neighbors, aggregate the exchanged models on their devices, and proceed with the next round of model training/fine-tuning.} 
    \label{fig:system_model}
    \vspace{-5mm}
\end{figure}

\textbf{4. Tailoring M3T FedFMs to Education Ecosystem:}
Here, we describe the system model envisioned for realizing a network of M3T FedFMs, as illustrated in Fig.~\ref{fig:system_model}(a). 
The system model follows the conventional ``star topology"~\cite{wu2024topology} in FL setting which includes a \textit{global server} interacting with a set of \textit{clients},\footnote{Star topology refers to the usage of client-to-server links for model aggregation and broadcast.} each described as follows:

\begin{enumerate}
    \item \textbf{Global server}, which hosts a comprehensive M3T FedFM consisting of globally aggregated versions of each available component, including modality encoders, task-specific heads, backbone structures, and context-specific prompt tuners. This global server selectively broadcasts necessary model components to the clients.

    \item \textbf{Clients}, comprising of three groups: institutions, instructors and students. The clients receive relevant subsets of the model components according to the modalities and tasks involved in the operations at each group from the global server.
    For instance, an instructor involved in a course might receive modality encoders and task heads corresponding to video, text, and image data for curriculum design, feedback generation, and content visualization. Also, a student in the same course may be provided with components that support video, audio, and text modalities necessary for classroom transcription, conceptual visualization, and supplementary research tasks. These clients subsequently transmit their locally trained/fine-tuned model parameters directly back to the global server for aggregation.
\end{enumerate}

Note that, as depicted in Fig.~\ref{fig:system_model}, additional modifications, such as introducing or removing connections between the individual client groups and changing the style of model aggregations, can be explored (e.g., to enhance model performance, convergence speed, and resource efficiency).

\section*{Unique Applications of M3T FedFMs for Education}


In the following, we discuss how M3T FedFMs can advance the privacy, personalization, and equity dimensions of next-generation intelligent education systems. We also present a set of forward-looking examples to concretely demonstrate how these advancements can be realized in practice.


\subsection*{\underline{Dimension 1:} Privacy-Enhanced M3T Intelligence} \label{sec:privacy}
M3T FedFMs naturally address longstanding data privacy concerns in educational ML applications by transmitting only model or gradient parameters between clients and the server, rather than raw sensitive data. 
By mitigating data privacy risks, M3T FedFMs enable greater participation from privacy-conscious individuals and institutions whose strict policies on raw-data sharing would have prevented them from contributing to the model training upon relying on the centralized training/fine-tuning of M3T FMs. This enhanced participation contributes to the development of models that generalize across varied educational settings. Below are three examples illustrating how M3T FedFMs can enable privacy-aware intelligence across M3T educational applications:

\begin{itemize}[leftmargin=4mm]
\item \textbf{Student Activity Traces:} Future education systems may integrate smartphone-based learning companions and augmented reality (AR) headsets that passively collect contextual data such as study hours, geolocation patterns (e.g., time spent in libraries or study zones), ambient noise, or device interactions \cite{antonioli2014augmented, bower2014augmented, kleftodimos2023location}. These data sources span various modalities such as time-series logs, location metadata, ambient audio, and app usage sequences. Such data is often privacy-sensitive in nature as it can reveal private user information, and thus cannot be shared or transferred across the network and must remain local to its data collecting unit/device. 
M3T FedFMs can process such collected data across the smartphones and AR devices to support education-related downstream tasks such as predicting study burnout, recommending optimal study windows, or modeling learning motivation, all without exposing raw activity data to external servers.
\item \textbf{Mental Health Assistance:} With the rise of Internet-of-Things (IoT) wearable biosensors, such as electroencephalogram (EEG) headbands and smartwatches equipped with heart rate variability (HRV) sensors, digital learning assistants can unobtrusively collect physiological, behavioral, and emotional cues to assess student well-being \cite{xu2018review, kim2024validity, aranberri2022reducing}. When combined with other privacy-sensitive modalities such as speech samples, writing patterns, and facial micro-expressions captured during classroom interactions or reflective exercises, this geo-distributed multi-modal data becomes a valuable resource for mental health analytics. Specifically, leveraging the collaborative training of M3T FedFMs over this data, digital learning assistants can be equipped with models for downstream tasks such as stress detection, mood tracking, and early intervention for depression or anxiety, all without exposing raw student data or violating privacy norms.

\item \textbf{Student Learning Outcome Prediction:} Future classrooms are expected to increasingly incorporate ambient sensors and camera-equipped devices, such as AI-driven smartboards and virtual reality (VR) learning environments, to assess student engagement in real-time \cite{lin2024impact, grewe2023can}. In this technological realm, features such as gaze tracking, posture analysis, voice tone, and note-taking behavior, derived from privacy-sensitive modalities (e.g., audio and video), offer deep insights into cognitive and behavioral states. However, due to their sensitive nature, such data cannot be shared across institutions/classrooms. Here, M3T FedFMs provide a viable solution by enabling distributed, privacy-preserving model training across institutions/classrooms. This allows for the development of generalizable models capable of predicting attention span, learning progress, and knowledge retention without compromising student privacy.
\end{itemize}

\subsection*{\underline{Dimension 2:} Personalization of M3T Intelligence} \label{sec:personalization}

Personalization is a foundational pillar of effective education, reflecting the need to tailor the learning experience to the unique characteristics, preferences, and needs of various \textit{educational entities}, including students, instructors, and institutions.
M3T FedFMs, with their inherently modular and flexible architectures, are uniquely positioned to support model personalization. Specifically, 
the notion of \textit{personalization} within M3T FedFMs can be understood from two complementary perspectives: \textit{(1) Soft Personalization (local fine-tuning):} At the core of this perspective lies the ability to perform \textit{local fine-tuning}, allowing clients to personalize prediction or generation tasks based on their own data distributions and contextual nuances, such as behavioral patterns, cultural or linguistic backgrounds, and interaction styles. M3T FedFMs leverage attention-based mechanisms and adaptable modules (e.g., prompt tuners and adapters) to personalize outputs dynamically, without the need for retraining the entire global model. 
\textit{(2) Hard Personalization (Architectural/Component Adaptation):} In this approach, personalization is embedded in the model architecture itself. Specifically, each client (e.g., institution, instructor, or student) is served a version of the M3T FedFM that contains only the components relevant to their available data modalities and educational tasks, such as specific modality encoders (e.g., for audio or video) and task heads (e.g., for problem solving, code generation, or essay evaluation). This selective architectural deployment ensures efficiency and relevance while maintaining interoperability with the broader intelligent education ecosystem. In the following, we describe three examples of personalization across client groups (students, instructors, institutions):

\begin{itemize}[leftmargin=4mm]
    \item \textbf{Student Personalization:} Students require personalized learning in various contexts, including concept explanation and problem-solving support. For example, some learners may benefit from visual explanations (e.g., video demonstrations or interactive visualizations), whereas others require textual or verbal guidance (e.g., spoken explanations, interactive Q\&A sessions). M3T FedFMs can leverage multi-modal inputs such as handwritten notes (image modality), spoken questions (audio), and clickstream behaviors (event logs), to address these varied needs by locally fine-tuning models (e.g., via adapters) for personalized concept explanations, adaptive problem-solving assistance, and real-time feedback.
    
    \item \textbf{Instructor Personalization:} Personalization for instructors often revolves around customized assessment generation and curricular support, which can differ based on the taught subject matter (e.g., essays, presentations, coding tasks, or visual projects).
    M3T FedFMs facilitate instructor personalization by employing specialized task heads and adapters that efficiently generate these customized assessments, minimizing preparation time and enhancing instructional quality. For instance, a language arts teacher may require automated feedback systems that assess creativity and narrative flow, whereas a computer science instructor might rely on auto-generated problem sets with dynamic test cases.
    
    \item \textbf{Institution Personalization:} Institutions vary in curricular standards, target outcomes, and infrastructural capabilities. A vocational training center focused on mechanical skills may use video-based object manipulation tasks, while a liberal arts college may emphasize text analysis. M3T FedFMs support this diversity by allowing each institution to personalize the model’s architecture, activating only relevant modalities (e.g., image and video for one, text and speech for another), and updating specific components (e.g., adapters or task heads) to reflect their educational mission.
\end{itemize}

\subsection*{\underline{Dimension 3:} Equitable and Inclusive M3T Intelligence} \label{sec:equity}
While \textit{personalization} focuses on tailoring educational models to the individual preferences, behaviors, or needs of specific users (e.g., students, instructors, or institutions), equity and inclusivity emphasize \textit{system-wide fairness and representation} across diverse social, cultural, linguistic, and infrastructural contexts. More specifically, personalization ensures that each user receives an optimized experience; equity and inclusivity ensure that every type of user (regardless of region, resources, identity, or participation patterns) is fairly represented in the training and utility of AI models. In this context, M3T FedFMs offer a practical pathway toward fostering a more equitable and inclusive educational ecosystem by accommodating variations in both curricular content and hardware/computation infrastructure across diverse learners and institutions. Specifically, unlike centralized models that often reflect dominant languages, curricula, or well-resourced environments/institutions, M3T FedFMs empower geographically distributed institutions to collaboratively train M3T FMs using locally relevant data while maintaining data sovereignty. 
Below, we present three examples that highlight how equity and inclusivity are advanced through M3T FedFMs in education:

\begin{itemize}[leftmargin=4mm]
\item \textbf{Cultural and Linguistic Representation:} In distributed educational systems, curriculum content, language, and cultural references may vary significantly across regions. Centralized models often fail to capture this diversity, especially for languages belonging to low-resource communities or locally relevant subjects. M3T FedFMs enable institutions to train/fine-tune  local models using culturally specific data, such as textbooks in indigenous languages or region-specific historical texts. As a result, the global model becomes more representative of diverse educational needs and fosters inclusivity in its learned knowledge.

\item \textbf{Infrastructure/Hardware-Aware Participation:} Educational entities participating in M3T FedFM training or fine-tuning often differ widely in their computational resources. While some (e.g., universities or research centers) may possess high-performance servers capable of full-scale model training, others (e.g., individual students or small schools) may rely on resource-constrained devices such as smartphones or tablets. 
To accommodate such disparities in computation capabilities, M3T FedFMs support \textit{modular engagement}, 
allowing resource-constrained eduction entities to contribute via lightweight computations (e.g., training only task heads or prompts) or to perform inference using relevant pre-trained components. This flexibility ensures that both high-end and low-resource education entities can benefit from and contribute to the collective learning process.

\item \textbf{Gender Bias and Fairness Mitigation:}
Centralized training of M3T FMs often risks amplifying existing gender biases, particularly when the underlying data disproportionately represents one gender over others. Such imbalance can lead to models that perform better for overrepresented groups while exhibiting reduced accuracy, relevance, or responsiveness for underrepresented genders, ultimately reinforcing inequality in educational outcomes. M3T FedFMs offer a more equitable alternative compared to centralized M3T FM training/fine-tuning by enabling distributed, gender-diverse participation in model training. Institutions and users across different regions and demographics can contribute model/gradient parameters reflecting balanced or marginalized gender identities without compromising privacy. 

\end{itemize}

\section*{Challenges and Open Directions of Federated Foundational Models (FedFMs) in Education} \label{sec:challenges}

Despite their potential, M3T FedFMs face unique deployment challenges in education. Below, we formulate overarching research questions aimed at addressing them.

\subsection*{Inter-Institution Heterogeneous Privacy Regulations and Their Impact on Data Availability}
As educational institutions across the globe adopt AI-driven systems, the deployment of M3T FedFMs in practice will become increasingly constrained by diverse and evolving privacy regulations. Specifically, legal frameworks such as the General Data Protection Regulation (GDPR) in the European Union and the Family Educational Rights and Privacy Act (FERPA) in the United States impose different requirements on how sensitive educational data, such as video, voice recordings, and physiological signals, can be used or shared across the education systems. These jurisdictional differences introduce a significant barrier to uniform collaboration in model training across institutions.
In FL, where raw data remains local and only model/gradient parameters are shared, the variability in legal constraints manifests as DP budgets or encryption standards applied to local updates. For example, an institution in a stricter jurisdiction may be obligated to inject stronger DP noise into model updates derived from video or speech data, reducing their informativeness relative to updates from
regions with less strict regulations in parameter sharing.
As a result, the aggregation process in M3T FedFMs becomes non-trivial: updates now vary not only in content and modality but also in privacy-induced distortion levels.
This privacy heterogeneity is especially problematic in educational contexts, where certain tasks (e.g., affect recognition or engagement tracking) heavily rely on privacy-sensitive modalities. In particular, a naïve aggregation of differentially distorted model updates from clients can inadvertently amplify the influence of under-regulated clients while marginalizing updates from more privacy-conscious clients, leading to skewed global model behavior.

$\star$ Given the rising importance of privacy-preserving AI in education, and the limited study of regulation-aware model aggregation in multi-modal FL \cite{liu2024mutual} and the unexplored study of regulation-based privacy in M3T FedFMs, an urgent research question is raised:
\textit{How can trust-aware aggregation mechanisms be designed in M3T Federated Foundation Models to fairly and effectively integrate updates subject to heterogeneous privacy regulations, while preserving convergence, modality balance, and cross-jurisdictional equity in global model behavior?}

\subsection*{Modality-Specific Characteristics and Transmission Overhead} 
While the above-discussed jurisdictional differences constrain data handling policies across institutions, an orthogonal and equally critical dimension arises from the inherent privacy sensitivity and computational demands associated with \textit{different input modalities} themselves. Specifically, in educational settings, the multi-modal nature of M3T FedFMs introduces distinct privacy risks across different input streams. While modalities such as text logs or quiz responses are generally considered lower-risk, others such as eye gaze, facial expressions, EEG signals, or audio recordings carry higher privacy sensitivity due to their biometric nature and potential to reveal deeply personal information. As ambient sensing technologies become more prevalent in classrooms, ensuring appropriate protection for these high-risk modalities is essential.
Compounding this challenge is the asymmetric contribution of modalities to different downstream educational tasks supported by M3T FedFMs. For example, facial expressions and vocal tone might be pivotal for engagement estimation, whereas textual responses are more relevant for concept mastery or personalized feedback. This variation makes uniform privacy-preserving strategies infeasible. Instead, techniques such as DP must be selectively applied based on each modality’s sensitivity and its utility for specific learning objectives. 


$\star$ Given that privacy calibration across modalities remains underexplored in M3T FedFMs, this raises a key open research question:
\textit{How can privacy-preserving techniques in M3T FedFMs be dynamically adapted across modalities to balance privacy risks and task-specific utility, especially when different modalities contribute asymmetrically to various tasks?}


\subsection*{User-Initiated Data Removal and the Need for Federated Unlearning} 
A critical challenge in privacy-aware educational systems is enabling users and institutions to revoke their data contributions after participation, an increasingly important right under regulations such as GDPR and FERPA. In the context of M3T FedFMs, this necessitates the development of effective \textit{federated unlearning} mechanisms: methods that can selectively remove the influence of a client’s data from the global model without requiring model retraining from scratch.
Unlike traditional centralized models, M3T FedFMs present unique obstacles for unlearning due to their modular structure, multi-modal data inputs, and decentralized training process. Specifically, client contributions are distributed across various components, such as modality encoders, adapters, and task heads, making their influence deeply entangled within the global model parameters. This makes it difficult to (1) accurately isolate and remove a client’s impact, and (2) maintain the global model’s utility, adaptation capability, and fairness for the remaining participants.

$\star$ While federated unlearning has begun to receive attention in classical FL settings \cite{halimi2022federated,liu2024survey}, it remains entirely unexplored in the context of M3T FedFMs. This gap raises a crucial and timely research question:
\textit{How can we design scalable, component-aware unlearning techniques for M3T FedFMs that ensure efficient, verifiable removal of user-contributed knowledge, while preserving model performance, fairness, and adaptability across heterogeneous and privacy-sensitive educational environments?}


\subsection*{Continual Learning}
Educational systems are inherently dynamic: new subjects are introduced, pedagogical approaches evolve, institutional priorities shift, and the nature of data modalities continually changes. In such an environment, static M3T FedFMs can quickly become misaligned with emerging learning objectives or newly introduced input modalities, limiting their adaptability and long-term relevance. 
This issue is amplified in federated/distributed settings, where decentralized and asynchronous model updates from various clients can complicate continual model adaptation. 
A central challenge arising from this distributed evolution is \textit{federated catastrophic forgetting}: as local model updates from clients are integrated sequentially or asynchronously, newly learned patterns (often specific to a subset of clients) can inadvertently overwrite previously acquired knowledge encoded in the global model. This forgetting effect is especially detrimental in education contexts, where preserving knowledge pertaining to personalized pedagogical methods (e.g., a student’s learning history or an institution’s domain-specific curriculum) is critical for long-term model effectiveness.
Specifically, without robust mechanisms to manage incremental learning and protect previously acquired knowledge, M3T FedFMs risk deteriorating in performance over time, especially for clients whose data distributions are no longer active but remain pedagogically important. This compromises the global model's reliability and generalization, reducing its value over time.

$\star$ Given that continual learning remains an emerging topic in the FM literature~\cite{ostapenko2022continual,yi2023towards,yang2025recent}, and is highly unexplored in the context of M3T FedFMs, this raises a pressing research question:
\textit{How can continual learning strategies for M3T FedFMs be designed to balance asynchronous client updates with prior knowledge retention, effectively mitigating federated catastrophic forgetting while supporting evolving educational tasks and modalities?}

\subsection*{Model Interpretability}
In educational settings, transparency and trust are paramount. Specifically, AI models that influence high-stakes decisions, such as grading, personalized feedback, skill assessment, or behavioral monitoring, must be explainable/interpretable to a wide range of stakeholders, including students, instructors, administrators, and parents. When the reasoning behind model outputs is unclear or opaque, it can undermine confidence, hinder adoption, and raise critical ethical concerns.
Interpretability becomes especially challenging in the context of M3T FedFMs, which introduce a compounded layer of complexity. First, their multi-modal nature involves inputs such as text, audio, video, and physiological signals, each varying in semantics, structure, and abstraction. Second, their modular architecture, comprising independently functioning components, such as prompt tuners, adapters, and task heads, makes it difficult to attribute predictions to specific modules or modalities. Third, the federated training paradigm adds further opacity: models are updated across decentralized clients with non-IID data distributions, meaning that the global model’s behavior emerges from a combination of locally trained, heterogeneous data sources. As a result, interpretability tools applied to the global model may fail to capture client-specific nuances or may produce misleading explanations when generalized across participants.
Together, compared to traditional centralized ML models, these challenges make it substantially harder to trace how specific input modalities or data features influence a given prediction, to identify hidden biases, or to justify model decisions for each separate client.

$\star$ Given that interpretability is already a nascent area in FMs~\cite{chen2022interpretable,rajendran2024learning,fu2024championing}, and remains almost unexplored in M3T FedFMs, this raises a foundational open research question: \textit{How can we design inherently interpretable M3T FedFMs that not only provide accurate outputs but also generate actionable, role-sensitive explanations aligned with the pedagogical and ethical demands of education systems?}








 

\section*{Conclusion}
In this position paper, we examined the emerging convergence of federated learning and foundation models within the education domain, framing the concept of \textit{multi-modal, multi-task federated foundation models (M3T FedFMs)} as a transformative step toward next-generation intelligent educational systems. We outlined the architectural structure of M3T FedFMs and discussed how their modular and distributed design offers a framework to address core needs in education, specifically: preserving privacy in learning processes, enabling model personalization, and promoting equity and inclusivity.
We also identified a set of open challenges and articulated key research questions designed to guide future inquiry in this nascent area.

\bibliography{sample}

\begin{thebibliography}{10}
\urlstyle{rm}
\expandafter\ifx\csname url\endcsname\relax
  \def\url#1{\texttt{#1}}\fi
\expandafter\ifx\csname urlprefix\endcsname\relax\def\urlprefix{URL }\fi
\expandafter\ifx\csname doiprefix\endcsname\relax\def\doiprefix{DOI: }\fi
\providecommand{\bibinfo}[2]{#2}
\providecommand{\eprint}[2][]{\url{#2}}

\bibitem{ofori2020using}
\bibinfo{author}{Ofori, F.}, \bibinfo{author}{Maina, E.} \&
  \bibinfo{author}{Gitonga, R.}
\newblock \bibinfo{journal}{\bibinfo{title}{Using machine learning algorithms
  to predict students’ performance and improve learning outcome: A literature
  based review}}.
\newblock {\emph{\JournalTitle{Journal of Information and Technology}}}
  \textbf{\bibinfo{volume}{4}}, \bibinfo{pages}{33--55} (\bibinfo{year}{2020}).

\bibitem{hridi2025privacy}
\bibinfo{author}{Hridi, A.~P.} \emph{et~al.}
\newblock \bibinfo{title}{Privacy-preserving distributed link predictions among
  peers in online classrooms using federated learning}.
\newblock In \emph{\bibinfo{booktitle}{Conference on Educational Data Mining
  (EDM)}} (\bibinfo{year}{2025}).

\bibitem{barua2022artificial}
\bibinfo{author}{Barua, P.~D.} \emph{et~al.}
\newblock \bibinfo{journal}{\bibinfo{title}{Artificial intelligence enabled
  personalised assistive tools to enhance education of children with
  neurodevelopmental disorders—a review}}.
\newblock {\emph{\JournalTitle{International Journal of Environmental Research
  and Public Health}}} \textbf{\bibinfo{volume}{19}}, \bibinfo{pages}{1192}
  (\bibinfo{year}{2022}).

\bibitem{ball2019applying}
\bibinfo{author}{Ball, R.} \emph{et~al.}
\newblock \bibinfo{title}{Applying machine learning to improve curriculum
  design}.
\newblock In \emph{\bibinfo{booktitle}{Proceedings of the 50th ACM Technical
  Symposium on Computer Science Education}}, \bibinfo{pages}{787--793}
  (\bibinfo{year}{2019}).

\bibitem{ebrahimi2025transition}
\bibinfo{author}{Ebrahimi, M.}, \bibinfo{author}{Sahay, R.},
  \bibinfo{author}{Hosseinalipour, S.} \& \bibinfo{author}{Akram, B.}
\newblock \bibinfo{journal}{\bibinfo{title}{The transition from centralized
  machine learning to federated learning for mental health in education: A
  survey of current methods and future directions}}.
\newblock {\emph{\JournalTitle{arXiv preprint arXiv:2501.11714}}}
  (\bibinfo{year}{2025}).

\bibitem{ingkavara2022use}
\bibinfo{author}{Ingkavara, T.}, \bibinfo{author}{Panjaburee, P.},
  \bibinfo{author}{Srisawasdi, N.} \& \bibinfo{author}{Sajjapanroj, S.}
\newblock \bibinfo{journal}{\bibinfo{title}{The use of a personalized learning
  approach to implementing self-regulated online learning}}.
\newblock {\emph{\JournalTitle{Computers and Education: Artificial
  Intelligence}}} \textbf{\bibinfo{volume}{3}}, \bibinfo{pages}{100086}
  (\bibinfo{year}{2022}).

\bibitem{xie2025review}
\bibinfo{author}{Xie, Y.}, \bibinfo{author}{Yang, L.}, \bibinfo{author}{Zhang,
  M.}, \bibinfo{author}{Chen, S.} \& \bibinfo{author}{Li, J.}
\newblock \bibinfo{journal}{\bibinfo{title}{A review of multimodal interaction
  in remote education: Technologies, applications, and challenges}}.
\newblock {\emph{\JournalTitle{Applied Sciences}}}
  \textbf{\bibinfo{volume}{15}}, \bibinfo{pages}{3937} (\bibinfo{year}{2025}).

\bibitem{griol2014developing}
\bibinfo{author}{Griol, D.}, \bibinfo{author}{Molina, J.~M.} \&
  \bibinfo{author}{De~Miguel, A.~S.}
\newblock \bibinfo{journal}{\bibinfo{title}{Developing multimodal
  conversational agents for an enhanced e-learning experience}}.
\newblock {\emph{\JournalTitle{ADCAIJ: Advances in Distributed Computing and
  Artificial Intelligence Journal}}} \textbf{\bibinfo{volume}{3}},
  \bibinfo{pages}{13--26} (\bibinfo{year}{2014}).

\bibitem{an2022no}
\bibinfo{author}{An, S.}, \bibinfo{author}{Kim, J.}, \bibinfo{author}{Kim, M.}
  \& \bibinfo{author}{Park, J.}
\newblock \bibinfo{title}{No task left behind: Multi-task learning of knowledge
  tracing and option tracing for better student assessment}.
\newblock In \emph{\bibinfo{booktitle}{Proceedings of the AAAI conference on
  artificial intelligence}}, vol.~\bibinfo{volume}{36},
  \bibinfo{pages}{4424--4431} (\bibinfo{year}{2022}).

\bibitem{geden2020predictive}
\bibinfo{author}{Geden, M.}, \bibinfo{author}{Emerson, A.},
  \bibinfo{author}{Rowe, J.}, \bibinfo{author}{Azevedo, R.} \&
  \bibinfo{author}{Lester, J.}
\newblock \bibinfo{title}{Predictive student modeling in educational games with
  multi-task learning}.
\newblock In \emph{\bibinfo{booktitle}{Proceedings of the AAAI Conference on
  Artificial Intelligence}}, vol.~\bibinfo{volume}{34},
  \bibinfo{pages}{654--661} (\bibinfo{year}{2020}).

\bibitem{kuchemann2025opportunities}
\bibinfo{author}{K{\"u}chemann, S.} \emph{et~al.}
\newblock \bibinfo{journal}{\bibinfo{title}{On opportunities and challenges of
  large multimodal foundation models in education}}.
\newblock {\emph{\JournalTitle{npj Science of Learning}}}
  \textbf{\bibinfo{volume}{10}}, \bibinfo{pages}{11} (\bibinfo{year}{2025}).

\bibitem{xu2024foundation}
\bibinfo{author}{Xu, T.} \emph{et~al.}
\newblock \bibinfo{journal}{\bibinfo{title}{Foundation models for education:
  Promises and prospects}}.
\newblock {\emph{\JournalTitle{IEEE Intelligent Systems}}}
  \textbf{\bibinfo{volume}{39}}, \bibinfo{pages}{20--24}
  (\bibinfo{year}{2024}).

\bibitem{gpt3}
\bibinfo{author}{Brown, T.~B.} \emph{et~al.}
\newblock \bibinfo{journal}{\bibinfo{title}{Language models are few-shot
  learners}}.
\newblock {\emph{\JournalTitle{Advances in neural information processing
  systems}}} \textbf{\bibinfo{volume}{33}}, \bibinfo{pages}{1877--1901}
  (\bibinfo{year}{2020}).

\bibitem{devlin2019bert}
\bibinfo{author}{Devlin, J.}, \bibinfo{author}{Chang, M.-W.},
  \bibinfo{author}{Lee, K.} \& \bibinfo{author}{Toutanova, K.}
\newblock \bibinfo{journal}{\bibinfo{title}{Bert: Pre-training of deep
  bidirectional transformers for language understanding}}.
\newblock {\emph{\JournalTitle{arXiv preprint arXiv:1810.04805}}}
  (\bibinfo{year}{2019}).

\bibitem{touvron2023llama}
\bibinfo{author}{Touvron, H.} \emph{et~al.}
\newblock \bibinfo{title}{Llama: Open and efficient foundation language models}
  (\bibinfo{year}{2023}).
\newblock \eprint{2302.13971}.

\bibitem{chowdhery2023palm}
\bibinfo{author}{Chowdhery, A.} \emph{et~al.}
\newblock \bibinfo{journal}{\bibinfo{title}{Palm: Scaling language modeling
  with pathways}}.
\newblock {\emph{\JournalTitle{Journal of Machine Learning Research}}}
  \textbf{\bibinfo{volume}{24}}, \bibinfo{pages}{1--113}
  (\bibinfo{year}{2023}).

\bibitem{openai2023gpt4}
\bibinfo{author}{OpenAI}.
\newblock \bibinfo{title}{Gpt-4 technical report}.
\newblock \bibinfo{howpublished}{\url{https://openai.com/research/gpt-4}}
  (\bibinfo{year}{2023}).
\newblock \bibinfo{note}{Accessed: 2025-06-15}.

\bibitem{team2023gemini}
\bibinfo{author}{Team, G.} \emph{et~al.}
\newblock \bibinfo{journal}{\bibinfo{title}{Gemini: a family of highly capable
  multimodal models}}.
\newblock {\emph{\JournalTitle{arXiv preprint arXiv:2312.11805}}}
  (\bibinfo{year}{2023}).

\bibitem{grattafiori2024llama}
\bibinfo{author}{Grattafiori, A.} \emph{et~al.}
\newblock \bibinfo{journal}{\bibinfo{title}{The llama 3 herd of models}}.
\newblock {\emph{\JournalTitle{arXiv preprint arXiv:2407.21783}}}
  (\bibinfo{year}{2024}).

\bibitem{guo2025human}
\bibinfo{author}{Guo, Y.} \emph{et~al.}
\newblock \bibinfo{journal}{\bibinfo{title}{Human-centric evaluation for
  foundation models}}.
\newblock {\emph{\JournalTitle{arXiv preprint arXiv:2506.01793}}}
  (\bibinfo{year}{2025}).

\bibitem{jia2024assessing}
\bibinfo{author}{Jia, Q.} \emph{et~al.}
\newblock \bibinfo{title}{On assessing the faithfulness of llm-generated
  feedback on student assignments}.
\newblock In \emph{\bibinfo{booktitle}{Proceedings of the 17th International
  Conference on Educational Data Mining}}, \bibinfo{pages}{491--499}
  (\bibinfo{year}{2024}).

\bibitem{mitra2024retllm}
\bibinfo{author}{Mitra, C.} \emph{et~al.}
\newblock \bibinfo{title}{Retllm-e: retrieval-prompt strategy for
  question-answering on student discussion forums}.
\newblock In \emph{\bibinfo{booktitle}{Proceedings of the AAAI Conference on
  Artificial Intelligence}}, vol.~\bibinfo{volume}{38},
  \bibinfo{pages}{23215--23223} (\bibinfo{year}{2024}).

\bibitem{molina2024leveraging}
\bibinfo{author}{Molina, I.~V.}, \bibinfo{author}{Montalvo, A.},
  \bibinfo{author}{Ochoa, B.}, \bibinfo{author}{Denny, P.} \&
  \bibinfo{author}{Porter, L.}
\newblock \bibinfo{journal}{\bibinfo{title}{Leveraging llm tutoring systems for
  non-native english speakers in introductory cs courses}}.
\newblock {\emph{\JournalTitle{arXiv preprint arXiv:2411.02725}}}
  (\bibinfo{year}{2024}).

\bibitem{zeide2018learner}
\bibinfo{author}{Zeide, E.} \& \bibinfo{author}{Nissenbaum, H.}
\newblock \bibinfo{journal}{\bibinfo{title}{Learner privacy in moocs and
  virtual education}}.
\newblock {\emph{\JournalTitle{Theory and Research in Education}}}
  \textbf{\bibinfo{volume}{16}}, \bibinfo{pages}{280--307}
  (\bibinfo{year}{2018}).

\bibitem{prinsloo2018student}
\bibinfo{author}{Prinsloo, P.} \& \bibinfo{author}{Slade, S.}
\newblock \bibinfo{title}{Student consent in learning analytics: the devil in
  the details?}
\newblock In \emph{\bibinfo{booktitle}{Learning analytics in higher
  education}}, \bibinfo{pages}{118--139} (\bibinfo{publisher}{Routledge},
  \bibinfo{year}{2018}).

\bibitem{mcmahan2017communication}
\bibinfo{author}{McMahan, B.}, \bibinfo{author}{Moore, E.},
  \bibinfo{author}{Ramage, D.}, \bibinfo{author}{Hampson, S.} \&
  \bibinfo{author}{y~Arcas, B.~A.}
\newblock \bibinfo{title}{Communication-efficient learning of deep networks
  from decentralized data}.
\newblock In \emph{\bibinfo{booktitle}{Artificial intelligence and
  statistics}}, \bibinfo{pages}{1273--1282} (\bibinfo{organization}{PMLR},
  \bibinfo{year}{2017}).

\bibitem{chen2022practical}
\bibinfo{author}{Chen, C.}, \bibinfo{author}{Lyu, L.}, \bibinfo{author}{Yu, H.}
  \& \bibinfo{author}{Chen, G.}
\newblock \bibinfo{journal}{\bibinfo{title}{Practical attribute reconstruction
  attack against federated learning}}.
\newblock {\emph{\JournalTitle{IEEE Transactions on Big Data}}}
  (\bibinfo{year}{2022}).

\bibitem{li2022ressfl}
\bibinfo{author}{Li, J.} \emph{et~al.}
\newblock \bibinfo{title}{Ressfl: A resistance transfer framework for defending
  model inversion attack in split federated learning}.
\newblock In \emph{\bibinfo{booktitle}{Proceedings of the IEEE/CVF conference
  on computer vision and pattern recognition}}, \bibinfo{pages}{10194--10202}
  (\bibinfo{year}{2022}).

\bibitem{el2022differential}
\bibinfo{author}{El~Ouadrhiri, A.} \& \bibinfo{author}{Abdelhadi, A.}
\newblock \bibinfo{journal}{\bibinfo{title}{Differential privacy for deep and
  federated learning: A survey}}.
\newblock {\emph{\JournalTitle{IEEE access}}} \textbf{\bibinfo{volume}{10}},
  \bibinfo{pages}{22359--22380} (\bibinfo{year}{2022}).

\bibitem{fang2021privacy}
\bibinfo{author}{Fang, H.} \& \bibinfo{author}{Qian, Q.}
\newblock \bibinfo{journal}{\bibinfo{title}{Privacy preserving machine learning
  with homomorphic encryption and federated learning}}.
\newblock {\emph{\JournalTitle{Future Internet}}}
  \textbf{\bibinfo{volume}{13}}, \bibinfo{pages}{94} (\bibinfo{year}{2021}).

\bibitem{chang2023privacy}
\bibinfo{author}{Chang, Y.}, \bibinfo{author}{Zhang, K.},
  \bibinfo{author}{Gong, J.} \& \bibinfo{author}{Qian, H.}
\newblock \bibinfo{journal}{\bibinfo{title}{Privacy-preserving federated
  learning via functional encryption, revisited}}.
\newblock {\emph{\JournalTitle{IEEE Transactions on Information Forensics and
  Security}}} \textbf{\bibinfo{volume}{18}}, \bibinfo{pages}{1855--1869}
  (\bibinfo{year}{2023}).

\bibitem{fachola2023federated}
\bibinfo{author}{Fachola, C.} \emph{et~al.}
\newblock \bibinfo{journal}{\bibinfo{title}{Federated learning for data
  analytics in education}}.
\newblock {\emph{\JournalTitle{Data}}} \textbf{\bibinfo{volume}{8}},
  \bibinfo{pages}{43} (\bibinfo{year}{2023}).

\bibitem{guo2020pedagogical}
\bibinfo{author}{Guo, S.}, \bibinfo{author}{Zeng, D.}, \bibinfo{author}{Dong,
  S.} \emph{et~al.}
\newblock \bibinfo{journal}{\bibinfo{title}{Pedagogical data analysis via
  federated learning toward education 4.0}}.
\newblock {\emph{\JournalTitle{American Journal of Education and Information
  Technology}}} \textbf{\bibinfo{volume}{4}}, \bibinfo{pages}{56--65}
  (\bibinfo{year}{2020}).

\bibitem{hridi2024revolutionizing}
\bibinfo{author}{Hridi, A.~P.}, \bibinfo{author}{Sahay, R.},
  \bibinfo{author}{Hosseinalipour, S.} \& \bibinfo{author}{Akram, B.}
\newblock \bibinfo{title}{Revolutionizing {AI}-assisted education with
  federated learning: A pathway to distributed, privacy-preserving, and
  debiased learning ecosystems}.
\newblock In \emph{\bibinfo{booktitle}{Proceedings of the AAAI Symposium
  Series}}, vol.~\bibinfo{volume}{3}, \bibinfo{pages}{297--303}
  (\bibinfo{year}{2024}).

\bibitem{chu2022mitigating}
\bibinfo{author}{Chu, Y.-W.} \emph{et~al.}
\newblock \bibinfo{title}{Mitigating biases in student performance prediction
  via attention-based personalized federated learning}.
\newblock In \emph{\bibinfo{booktitle}{Proceedings of the 31st ACM
  International Conference on Information \& Knowledge Management}},
  \bibinfo{pages}{3033--3042} (\bibinfo{year}{2022}).

\bibitem{chu2024multi}
\bibinfo{author}{Chu, Y.-W.} \emph{et~al.}
\newblock \bibinfo{journal}{\bibinfo{title}{Multi-layer personalized federated
  learning for mitigating biases in student predictive analytics}}.
\newblock {\emph{\JournalTitle{IEEE Transactions on Emerging Topics in
  Computing}}}  (\bibinfo{year}{2024}).

\bibitem{borazjani2025multi}
\bibinfo{author}{Borazjani, K.} \emph{et~al.}
\newblock \bibinfo{journal}{\bibinfo{title}{Multi-modal multi-task ({M3T})
  federated foundation models for embodied {AI}: Potentials and challenges for
  edge integration}}.
\newblock {\emph{\JournalTitle{arXiv preprint arXiv:2505.11191}}}
  (\bibinfo{year}{2025}).

\bibitem{nadimi2025multi}
\bibinfo{author}{Nadimi, F.}, \bibinfo{author}{Abdisarabshali, P.},
  \bibinfo{author}{Borazjani, K.}, \bibinfo{author}{Chakareski, J.} \&
  \bibinfo{author}{Hosseinalipour, S.}
\newblock \bibinfo{journal}{\bibinfo{title}{Multi-modal multi-task federated
  foundation models for next-generation extended reality systems: Towards
  privacy-preserving distributed intelligence in {AR/VR/MR}}}.
\newblock {\emph{\JournalTitle{arXiv preprint arXiv:2506.05683}}}
  (\bibinfo{year}{2025}).

\bibitem{chen2024disentanglement}
\bibinfo{author}{Chen, J.} \& \bibinfo{author}{Zhang, A.}
\newblock \bibinfo{title}{On disentanglement of asymmetrical knowledge transfer
  for modality-task agnostic federated learning}.
\newblock In \emph{\bibinfo{booktitle}{Proceedings of the AAAI Conference on
  Artificial Intelligence}}, vol.~\bibinfo{volume}{38},
  \bibinfo{pages}{11311--11319} (\bibinfo{year}{2024}).

\bibitem{guo2024scattering}
\bibinfo{author}{Guo, W.}, \bibinfo{author}{Li, S.} \& \bibinfo{author}{Yang,
  J.}
\newblock \bibinfo{title}{Scattering prompt tuning: A fine-tuned foundation
  model for sar object recognition}.
\newblock In \emph{\bibinfo{booktitle}{Proceedings of the IEEE/CVF Conference
  on Computer Vision and Pattern Recognition}}, \bibinfo{pages}{3056--3065}
  (\bibinfo{year}{2024}).

\bibitem{jia2022visual}
\bibinfo{author}{Jia, M.} \emph{et~al.}
\newblock \bibinfo{title}{Visual prompt tuning}.
\newblock In \emph{\bibinfo{booktitle}{European conference on computer
  vision}}, \bibinfo{pages}{709--727} (\bibinfo{organization}{Springer},
  \bibinfo{year}{2022}).

\bibitem{long2024dual}
\bibinfo{author}{Long, G.}, \bibinfo{author}{Shen, T.}, \bibinfo{author}{Jiang,
  J.}, \bibinfo{author}{Blumenstein, M.} \emph{et~al.}
\newblock \bibinfo{journal}{\bibinfo{title}{Dual-personalizing adapter for
  federated foundation models}}.
\newblock {\emph{\JournalTitle{Advances in Neural Information Processing
  Systems}}} \textbf{\bibinfo{volume}{37}}, \bibinfo{pages}{39409--39433}
  (\bibinfo{year}{2024}).

\bibitem{zhang2022contrastive}
\bibinfo{author}{Zhang, M.} \& \bibinfo{author}{R{\'e}, C.}
\newblock \bibinfo{journal}{\bibinfo{title}{Contrastive adapters for foundation
  model group robustness}}.
\newblock {\emph{\JournalTitle{Advances in Neural Information Processing
  Systems}}} \textbf{\bibinfo{volume}{35}}, \bibinfo{pages}{21682--21697}
  (\bibinfo{year}{2022}).

\bibitem{yang2024low}
\bibinfo{author}{Yang, M.} \emph{et~al.}
\newblock \bibinfo{journal}{\bibinfo{title}{Low-rank adaptation for foundation
  models: A comprehensive review}}.
\newblock {\emph{\JournalTitle{arXiv preprint arXiv:2501.00365}}}
  (\bibinfo{year}{2024}).

\bibitem{wen2023batched}
\bibinfo{author}{Wen, Y.} \& \bibinfo{author}{Chaudhuri, S.}
\newblock \bibinfo{journal}{\bibinfo{title}{Batched low-rank adaptation of
  foundation models}}.
\newblock {\emph{\JournalTitle{arXiv preprint arXiv:2312.05677}}}
  (\bibinfo{year}{2023}).

\bibitem{chen2024feddat}
\bibinfo{author}{Chen, H.}, \bibinfo{author}{Zhang, Y.},
  \bibinfo{author}{Krompass, D.}, \bibinfo{author}{Gu, J.} \&
  \bibinfo{author}{Tresp, V.}
\newblock \bibinfo{title}{Feddat: An approach for foundation model finetuning
  in multi-modal heterogeneous federated learning}.
\newblock In \emph{\bibinfo{booktitle}{Proceedings of the AAAI Conference on
  Artificial Intelligence}}, vol.~\bibinfo{volume}{38},
  \bibinfo{pages}{11285--11293} (\bibinfo{year}{2024}).

\bibitem{borazjani2025redefining}
\bibinfo{author}{Borazjani, K.}, \bibinfo{author}{Abdisarabshali, P.},
  \bibinfo{author}{Khosravan, N.} \& \bibinfo{author}{Hosseinalipour, S.}
\newblock \bibinfo{journal}{\bibinfo{title}{Redefining non-iid data in
  federated learning for computer vision tasks: Migrating from labels to
  embeddings for task-specific data distributions}}.
\newblock {\emph{\JournalTitle{arXiv preprint arXiv:2503.14553}}}
  (\bibinfo{year}{2025}).

\bibitem{borazjani2024multi}
\bibinfo{author}{Borazjani, K.}, \bibinfo{author}{Khosravan, N.},
  \bibinfo{author}{Ying, L.} \& \bibinfo{author}{Hosseinalipour, S.}
\newblock \bibinfo{journal}{\bibinfo{title}{Multi-modal federated learning for
  cancer staging over non-iid datasets with unbalanced modalities}}.
\newblock {\emph{\JournalTitle{IEEE Transactions on Medical Imaging}}}
  (\bibinfo{year}{2024}).

\bibitem{parasnis2023connectivity}
\bibinfo{author}{Parasnis, R.}, \bibinfo{author}{Hosseinalipour, S.},
  \bibinfo{author}{Chu, Y.-W.}, \bibinfo{author}{Chiang, M.} \&
  \bibinfo{author}{Brinton, C.~G.}
\newblock \bibinfo{title}{Connectivity-aware semi-decentralized federated
  learning over time-varying d2d networks}.
\newblock In \emph{\bibinfo{booktitle}{Proceedings of the Twenty-fourth
  International Symposium on Theory, Algorithmic Foundations, and Protocol
  Design for Mobile Networks and Mobile Computing}}, \bibinfo{pages}{31--40}
  (\bibinfo{year}{2023}).

\bibitem{chai2019towards}
\bibinfo{author}{Chai, Z.} \emph{et~al.}
\newblock \bibinfo{title}{Towards taming the resource and data heterogeneity in
  federated learning}.
\newblock In \emph{\bibinfo{booktitle}{2019 USENIX conference on operational
  machine learning (OpML 19)}}, \bibinfo{pages}{19--21} (\bibinfo{year}{2019}).

\bibitem{li2025open}
\bibinfo{author}{Li, X.}, \bibinfo{author}{Peng, L.}, \bibinfo{author}{Wang,
  Y.-P.} \& \bibinfo{author}{Zhang, W.}
\newblock \bibinfo{journal}{\bibinfo{title}{Open challenges and opportunities
  in federated foundation models towards biomedical healthcare}}.
\newblock {\emph{\JournalTitle{BioData Mining}}} \textbf{\bibinfo{volume}{18}},
  \bibinfo{pages}{2} (\bibinfo{year}{2025}).

\bibitem{abdisarabshali2025hierarchical}
\bibinfo{author}{Abdisarabshali, P.} \emph{et~al.}
\newblock \bibinfo{journal}{\bibinfo{title}{Hierarchical federated foundation
  models over wireless networks for multi-modal multi-task intelligence:
  Integration of edge learning with {D2D/P2P}-enabled fog learning
  architectures}}.
\newblock {\emph{\JournalTitle{arXiv preprint arXiv:2509.03695}}}
  (\bibinfo{year}{2025}).

\bibitem{wu2024topology}
\bibinfo{author}{Wu, J.} \emph{et~al.}
\newblock \bibinfo{journal}{\bibinfo{title}{Topology-aware federated learning
  in edge computing: A comprehensive survey}}.
\newblock {\emph{\JournalTitle{ACM Computing Surveys}}}
  \textbf{\bibinfo{volume}{56}}, \bibinfo{pages}{1--41} (\bibinfo{year}{2024}).

\bibitem{antonioli2014augmented}
\bibinfo{author}{Antonioli, M.}, \bibinfo{author}{Blake, C.} \&
  \bibinfo{author}{Sparks, K.}
\newblock \bibinfo{journal}{\bibinfo{title}{Augmented reality applications in
  education}}.
\newblock {\emph{\JournalTitle{The Journal of technology studies}}}
  \bibinfo{pages}{96--107} (\bibinfo{year}{2014}).

\bibitem{bower2014augmented}
\bibinfo{author}{Bower, M.}, \bibinfo{author}{Howe, C.},
  \bibinfo{author}{McCredie, N.}, \bibinfo{author}{Robinson, A.} \&
  \bibinfo{author}{Grover, D.}
\newblock \bibinfo{journal}{\bibinfo{title}{Augmented reality in
  education--cases, places and potentials}}.
\newblock {\emph{\JournalTitle{Educational Media International}}}
  \textbf{\bibinfo{volume}{51}}, \bibinfo{pages}{1--15} (\bibinfo{year}{2014}).

\bibitem{kleftodimos2023location}
\bibinfo{author}{Kleftodimos, A.}, \bibinfo{author}{Moustaka, M.} \&
  \bibinfo{author}{Evagelou, A.}
\newblock \bibinfo{journal}{\bibinfo{title}{Location-based augmented reality
  for cultural heritage education: Creating educational, gamified
  location-based ar applications for the prehistoric lake settlement of
  dispilio}}.
\newblock {\emph{\JournalTitle{Digital}}} \textbf{\bibinfo{volume}{3}},
  \bibinfo{pages}{18--45} (\bibinfo{year}{2023}).

\bibitem{xu2018review}
\bibinfo{author}{Xu, J.} \& \bibinfo{author}{Zhong, B.}
\newblock \bibinfo{journal}{\bibinfo{title}{Review on portable eeg technology
  in educational research}}.
\newblock {\emph{\JournalTitle{Computers in Human Behavior}}}
  \textbf{\bibinfo{volume}{81}}, \bibinfo{pages}{340--349}
  (\bibinfo{year}{2018}).

\bibitem{kim2024validity}
\bibinfo{author}{Kim, H.~J.}, \bibinfo{author}{Park, Y.} \&
  \bibinfo{author}{Lee, J.}
\newblock \bibinfo{journal}{\bibinfo{title}{The validity of heart rate
  variability (hrv) in educational research and a synthesis of
  recommendations}}.
\newblock {\emph{\JournalTitle{Educational Psychology Review}}}
  \textbf{\bibinfo{volume}{36}}, \bibinfo{pages}{42} (\bibinfo{year}{2024}).

\bibitem{aranberri2022reducing}
\bibinfo{author}{Aranberri-Ruiz, A.}, \bibinfo{author}{Aritzeta, A.},
  \bibinfo{author}{Olarza, A.}, \bibinfo{author}{Soroa, G.} \&
  \bibinfo{author}{Mindeguia, R.}
\newblock \bibinfo{journal}{\bibinfo{title}{Reducing anxiety and social stress
  in primary education: a breath-focused heart rate variability biofeedback
  intervention}}.
\newblock {\emph{\JournalTitle{International Journal of Environmental Research
  and Public Health}}} \textbf{\bibinfo{volume}{19}}, \bibinfo{pages}{10181}
  (\bibinfo{year}{2022}).

\bibitem{lin2024impact}
\bibinfo{author}{Lin, X.~P.}, \bibinfo{author}{Li, B.~B.},
  \bibinfo{author}{Yao, Z.~N.}, \bibinfo{author}{Yang, Z.} \&
  \bibinfo{author}{Zhang, M.}
\newblock \bibinfo{journal}{\bibinfo{title}{The impact of virtual reality on
  student engagement in the classroom--a critical review of the literature}}.
\newblock {\emph{\JournalTitle{Frontiers in Psychology}}}
  \textbf{\bibinfo{volume}{15}}, \bibinfo{pages}{1360574}
  (\bibinfo{year}{2024}).

\bibitem{grewe2023can}
\bibinfo{author}{Grewe, M.} \& \bibinfo{author}{Gie, L.}
\newblock \bibinfo{journal}{\bibinfo{title}{Can virtual reality have a positive
  influence on student engagement?}}
\newblock {\emph{\JournalTitle{South African Journal of Higher Education}}}
  \textbf{\bibinfo{volume}{37}}, \bibinfo{pages}{124--141}
  (\bibinfo{year}{2023}).

\bibitem{liu2024mutual}
\bibinfo{author}{Liu, X.}, \bibinfo{author}{Cai, S.}, \bibinfo{author}{He, R.}
  \& \bibinfo{author}{Yuan, J.}
\newblock \bibinfo{journal}{\bibinfo{title}{Mutual gradient inversion:
  Unveiling privacy risks of federated learning on multi-modal signals}}.
\newblock {\emph{\JournalTitle{IEEE Signal Processing Letters}}}
  (\bibinfo{year}{2024}).

\bibitem{halimi2022federated}
\bibinfo{author}{Halimi, A.}, \bibinfo{author}{Kadhe, S.},
  \bibinfo{author}{Rawat, A.} \& \bibinfo{author}{Baracaldo, N.}
\newblock \bibinfo{journal}{\bibinfo{title}{Federated unlearning: How to
  efficiently erase a client in fl?}}
\newblock {\emph{\JournalTitle{arXiv preprint arXiv:2207.05521}}}
  (\bibinfo{year}{2022}).

\bibitem{liu2024survey}
\bibinfo{author}{Liu, Z.} \emph{et~al.}
\newblock \bibinfo{journal}{\bibinfo{title}{A survey on federated unlearning:
  Challenges, methods, and future directions}}.
\newblock {\emph{\JournalTitle{ACM Computing Surveys}}}
  \textbf{\bibinfo{volume}{57}}, \bibinfo{pages}{1--38} (\bibinfo{year}{2024}).

\bibitem{ostapenko2022continual}
\bibinfo{author}{Ostapenko, O.} \emph{et~al.}
\newblock \bibinfo{title}{Continual learning with foundation models: An
  empirical study of latent replay}.
\newblock In \emph{\bibinfo{booktitle}{Conference on lifelong learning
  agents}}, \bibinfo{pages}{60--91} (\bibinfo{organization}{PMLR},
  \bibinfo{year}{2022}).

\bibitem{yi2023towards}
\bibinfo{author}{Yi, H.} \emph{et~al.}
\newblock \bibinfo{journal}{\bibinfo{title}{Towards general purpose medical ai:
  Continual learning medical foundation model}}.
\newblock {\emph{\JournalTitle{arXiv preprint arXiv:2303.06580}}}
  (\bibinfo{year}{2023}).

\bibitem{yang2025recent}
\bibinfo{author}{Yang, Y.} \emph{et~al.}
\newblock \bibinfo{journal}{\bibinfo{title}{Recent advances of foundation
  language models-based continual learning: A survey}}.
\newblock {\emph{\JournalTitle{ACM Computing Surveys}}}
  \textbf{\bibinfo{volume}{57}}, \bibinfo{pages}{1--38} (\bibinfo{year}{2025}).

\bibitem{chen2022interpretable}
\bibinfo{author}{Chen, J.} \emph{et~al.}
\newblock \bibinfo{journal}{\bibinfo{title}{Interpretable rna foundation model
  from unannotated data for highly accurate rna structure and function
  predictions}}.
\newblock {\emph{\JournalTitle{arXiv preprint arXiv:2204.00300}}}
  (\bibinfo{year}{2022}).

\bibitem{rajendran2024learning}
\bibinfo{author}{Rajendran, G.}, \bibinfo{author}{Buchholz, S.},
  \bibinfo{author}{Aragam, B.}, \bibinfo{author}{Sch{\"o}lkopf, B.} \&
  \bibinfo{author}{Ravikumar, P.}
\newblock \bibinfo{journal}{\bibinfo{title}{Learning interpretable concepts:
  Unifying causal representation learning and foundation models}}.
\newblock {\emph{\JournalTitle{arXiv preprint arXiv:2402.09236}}}
  (\bibinfo{year}{2024}).

\bibitem{fu2024championing}
\bibinfo{author}{Fu, S.}, \bibinfo{author}{Chen, Y.}, \bibinfo{author}{Wang,
  Y.} \& \bibinfo{author}{Tao, D.}
\newblock \bibinfo{journal}{\bibinfo{title}{On championing foundation models:
  From explainability to interpretability}}.
\newblock {\emph{\JournalTitle{arXiv preprint arXiv:2410.11444}}}
  (\bibinfo{year}{2024}).

\end{thebibliography}






\end{document}